\documentstyle[twocolumn]{article}  

\title{Centering in Japanese Discourse \\
       (APPEARED IN COLING90, Helsinki)}
\author{\begin{tabular}{ccc}
Marilyn Walker &  Masayo Iida  & Sharon Cote\\
University of Pennsylvania & Hewlett-Packard Laboratories & University
of Pennsylvania\\
Computer Science Dept.\thanks{This research was
partially funded by ARO grants
DAAG29-84-K-0061 and DAAL03-89-C0031PRI, DARPA grant N00014-85-K0018, and
NSF grant MCS-82-19196 at the University of Pennsylvania,
and by Hewlett-Packard Laboratories.} & Human-Computer Interaction
Dept. & Linguistics Dept.\\
Philadelphia, PA 19104 & Palo Alto, CA 94303 & Philadelphia, PA 19104  \\
lyn@linc.cis.upenn.edu & iida@csli.stanford.edu & cote@linc.cis.upenn.edu
\end{tabular}
}



\newcommand{\displaytitle}{\begin{center}
{\Large Centering in Japanese Discourse}\\[10mm]
(APPEARED IN COLING90, Helsinki) \\[5mm]
{\large
\begin{tabular}{ccc}
Marilyn Walker &  Masayo Iida  & Sharon Cote\\
University of Pennsylvania & Hewlett-Packard Laboratories & University
of Pennsylvania\\
Computer Science Dept.
& Human-Computer Interaction
Dept. & Linguistics Dept.\\
Philadelphia, PA 19104 & Palo Alto, CA 94303 & Philadelphia, PA 19104  \\
lyn@linc.cis.upenn.edu & iida@csli.stanford.edu & cote@linc.cis.upenn.edu
\end{tabular}}\\[10mm]
\end{center}}

\begin{document}           
\bibliographystyle{alpha}  

\twocolumn[\displaytitle]



\begin{abstract}
In this paper we propose a computational treatment of the resolution
of zero pronouns in Japanese discourse, using an adaptation of the
centering algorithm. We are able to factor language-specific
dependencies into one parameter of the centering algorithm.
Previous analyses have stipulated that a zero
pronoun and its cospecifier must share a grammatical function
property such as {\sc Subject} or {\sc NonSubject}.  We show that this
property-sharing stipulation is unneeded. In addition we propose the
notion of {\sc topic ambiguity} within the centering framework, which
predicts some ambiguities that occur in Japanese discourse.  This
analysis has implications for the design of language-independent
discourse modules for Natural Language systems. The centering
algorithm has been implemented in an HPSG Natural Language system with
both English and Japanese grammars.
\end{abstract}

\section{Introduction}
\label{intro-sec}

   Japanese is a language well-known for grammaticization of discourse
function. It is rich with ways for speakers to indicate the information
status of the discourse entities they are talking about.  Japanese allows a
speaker to clearly indicate topic-hood, along with the grammatical
functions such as subject, object and object2, by using the morphological
case markers {\em wa, ga, o, ni}.  In addition, it provides morphological
means to indicate speaker's perspective through the use of verbal
compounding, i.e.  the addition of suffixes such as {\em kureta, kita} (See
section 3).  Unexpressed arguments of the verb are common; these are
known as zero pronouns.

Because there are zero pronouns and because Japanese is a head-final
language with otherwise relatively free word order, there could, in
principle, be a great deal of ambiguity. However this is not the case.
Speakers are assumed to be cooperative, to be collaborating with the hearer
in conversation, and to be ensuring that each utterance is relevant and
coherent in the context of what was said before \cite{Grice75,SSJ74}. We
believe that speakers do not choose to express their thoughts through
arbitrary syntactic constructions, but that there is some correspondence
between choice of syntactic construction, what the speaker wants to convey,
and aspects of the current discourse situation\cite{Prince85}.

Within a theory of discourse, {\sc centering} is a computational model
of the process by which a speaker and hearer make obvious to one
another their assumptions about the salience of discourse entities.
Using pronominal referring expressions is one way for discourse
participants to do this. We propose that the resolution of zero
pronouns is constrained by centering, and ambiguity is thereby
reduced.

Centering has its computational foundations in the work of Grosz and
Sidner\cite{Grosz77,Sidner79,GS85} and was further developed by Grosz,
Joshi and Weinstein\cite{GJW83,GJW86,JW81}. It is formalized as a
system of constraints and rules, which can, as part of a computational
discourse model, act to control inferencing\cite{JW81}.  Brennan,
Friedman and Pollard use these rules and constraints to develop an
algorithm for resolving the co-specifiers of
pronouns\cite{BFP87,Walker89b}. Our analysis uses an adaptation of this
algorithm.  By making full use of the centering formalism,
we avoid the postulation of additional mechanisms, e.g. property
sharing\cite{Kameyama86a}.

In addition, we propose a notion of {\sc topic ambiguity}, which
characterizes some ambiguities in Japanese discourse that are allowed by
the centering process.  Topic ambiguity has been ignored in previous
accounts of Japanese zero pronoun resolution, but it explains the
availability of interpretations that previous accounts would predict as
ungrammatical. Centering gives us a computational way of determining when a
zero pronoun may be assigned {\sc Topic}.

This analysis informs the design of language independent discourse
processing modules for Natural Language systems.  We propose that the
centering component of a discourse processing module can be constructed in
a language independent fashion, up to the declaration of a language-specific
value for one variable in the algorithm, i.e., Cf list ranking (see
section \ref{cent-form}).  The centering algorithm has been
implemented in an HPSG Natural Language system with both English and
Japanese grammars.

\section{The Centering Formalism}
\label{cent-form}
The modeling of attentional state in discourse by
centering depends on analyzing each pair of utterances in a discourse
according to a set of transitions. These transitions are a measure of
the coherence of the segment of discourse in which the utterance
occurs.  Each utterance in a discourse has associated with it a set of
discourse entities called {\sc forward-looking centers}, $\rm{Cf}$,
and a special member of this set called the {\sc backward-looking
center}, $\rm{Cb}$.  The {\sc forward-looking centers} are ranked according
to discourse salience; the
highest ranked member of the set is the {\sc preferred center},
$\rm{Cp}$. With these definitions we can give the constraints:
\nopagebreak
\begin{itemize}
\item
{\bf CONSTRAINTS} \\
For each $\rm{U_{i}}$ in a discourse segment $\rm{U_{1}, \ldots ,U_{m}}$:
\vspace{-3.0ex}
   \begin{enumerate}
      \item There is precisely one $\rm{Cb}$.
      \item Every element of $\rm{Cf(U_{i}})$ must be realized\footnote
{An utterance U (of some phrase, not necessarily a full clause), {\em
realizes\/} {\bf c} if {\bf c} is an element of the situation
described by U, or {\bf c} is the semantic interpretation of some
subpart of U.}  in $\rm{U_{i}}$.
      \item The center, $\rm{Cb(U_{i})}$, is the highest-ranked element
of $\rm{Cf(U_{i-1})}$ that
                 is realized in $\rm{U_{i}}$.
    \end{enumerate}
\end{itemize}

The typology of transitions from one utterance, $\rm{U_{i}}$, to the next
is based on two factors: whether the backward-looking center, $\rm{Cb}$, is
the same from $\rm{U_{i-1}}$ to $\rm{U_{i}}$, and whether this discourse
entity is the same as the preferred center, $\rm{Cp}$ of $\rm{U_{i}}$.
Backward-looking centers are often pronominalized and discourses that
continue centering the same entity are more coherent than those that shift
from one center to another.  This means that some transitions are preferred
over others. These two facts give us the rules:

\begin{itemize}
\item
{\bf RULES} \\
For each $\rm{U_{i}}$ in a discourse segment $\rm{U_{1}, \ldots ,U_{m}}$:
\vspace{-3.0ex}
\nopagebreak
   \begin{enumerate}
     \item If some element of $\rm{Cf(U_{i-1})}$ is realized as a
pronoun in $\rm{U_{i}}$,
              then so is $\rm{Cb(U_{i})}$.
     \item Transition states are ordered.
   {\sc Continuing\/} is preferred to {\sc retaining\/} is preferred to
 {\sc shifting-1} is preferred to {\sc shifting}\footnote{\cite{BFP87}
introduces the distinction between SHIFTING-1 and SHIFTING.}.
   \end{enumerate}
\end{itemize}

The transition states that are used in the rules are defined in Figure
\ref{state-fig}, ({\sc backward-looking center} $ = \rm{Cb}$, {\sc
preferred Center} $ = \rm{Cp}$).

\begin{small}
\begin{figure}[htb]
\setlength{\unitlength}{.65in}
\begin{flushright}
\begin{picture}(4,2)

\put(0,0){\framebox(2,1){RETAINING}}
\put(0,1){\framebox(2,1){CONTINUING}}
\put(2,1){\framebox(2,1){SHIFTING-1}}
\put(2,0){\framebox(2,1){SHIFTING}}

\put(1,2.2){\makebox(0,0){$\rm{Cb(U_{i}) = Cb(U_{i-1}}$)}}
\put(3,2.2){\makebox(0,0){$\rm{Cb(U_{i}) \neq Cb(U_{i-1}}$)}}
\put(-.33,1.65){\makebox(0,0){$\rm{Cb(U_{i})}$}}
\put(-.25,1.45){\makebox(0,0){$=$}}
\put(-.33,1.25){\makebox(0,0){$\rm{Cp(U_{i})}$}}
\put(-.33,0.70){\makebox(0,0){$\rm{Cb(U_{i})}$}}
\put(-.25,0.5){\makebox(0,0){$\neq$}}
\put(-.33,0.30 ){\makebox(0,0){$\rm{Cp(U_{i})}$}}

\end{picture}
\end{flushright}
\normalsize

\caption{ Transition States}
\label{state-fig}
\end{figure}
\end{small}

The centering algorithm incorporates these rules and constraints in
addition to linguistic constraints on coreference\cite{BFP87}.  The
behavior of the centering algorithm for the resolution of pronouns is
largely determined by the ranking of the items on the forward center
list, $\rm{Cf}$, because, as per Constraint 3, this ranking determines
from among the elements that are realized in the next utterance, which
of them will be the $\rm{Cb}$ for that utterance.
Although all of the factors that contribute to the $\rm{Cf}$ ranking
have not been determined, syntax and lexical semantics have an
effect\cite{Prince81,Prince85,Hudson88,Brennan89,GJW86,JW81,BF83}.
We postulate that this ordering will vary from language to language
depending on the means the language provides for expressing discourse
functions.  Our adaptation of the algorithm for Japanese consists of
substituting a different ranking of the forward centers list
$\rm{Cf}$. In every other way, the algorithm functions exactly as it
is for English.

\section{Centering in Japanese}

In order to apply the centering algorithm to the resolution of zero
pronouns in Japanese, we must determine how to order the forward
centers list, $\rm{Cf}$.  The function {\sc topic} is indicated by the
morphological marker {\em wa}, along with {\sc subject} ({\em ga}),
{\sc object} ({\em o}), and {\sc object2} ({\em ni}).  The optional
use of {\em wa} picks out the most salient entity in the discourse.
In addition, Kuno proposed the notion of {\sc empathy}, which is the
perspective from which a speaker describes an event\cite{Kuno73}.  The
realization of speaker's empathy is especially important when
describing an event involving some transfer.  For example, there is no
way to describe a {\em giving\/} and {\em receiving\/} situation
objectively\cite{Kuno-Kab77}.  In (1), the use of the past tense {\em
kureta} of the verb {\em kureru\/}, indicates the speaker's empathy
with the discourse entity realized in object position\footnote{We
use identifiers of all capital letters to denote the discourse entity
realized by the corresponding string.  Centers are semantic entities,
not syntactic ones.}.

\noindent (1)\\
\footnotesize
\begin{tabular}{llllll}
Hanako wa & Taroo ni & hon o & kureta. & & \\
top-subj & obj2 & book obj & give-past & & \\
\multicolumn{6}{l}{{\it ``Hanako gave Taroo a book.''}}\\
\multicolumn{6}{l}{EMPATHY=OBJ2=TAROO} \\
\end{tabular}
\normalsize

In (2), the speaker's empathy with the subject entity's
perspective is indicated using {\em yatta}, the past tense of the verb
{\em yaru\/}.

\noindent (2)\\
\footnotesize
\begin{tabular}{llllll}
Hanako wa & Taroo ni & hon o & yatta. & & \\
top-subj & obj2 & book obj & give-past & & \\
\multicolumn{6}{l}{{\it ``Hanako gave Taroo a book.''}}\\
\multicolumn{6}{l}{EMPATHY=SUBJ=HANAKO} \\
\end{tabular}
\normalsize

The use of deictic verbs such as {\it kuru\/} (`come'), and {\it
iku\/} (`go') also indicate speaker's perspective.

Kuno calls a verb that is sensitive to the speaker's perspective an
{\sc Empathy-loaded} verb, and defines {\sc Empathy locus} as the
argument position whose referent the speaker is identifying
with\footnote{The speaker does not necessarily take his/her own
perspective to describe an event in which s/he is involved.}.
Any Japanese verb can be made into an empathy-loaded verb by using an
empathy-loaded verb as an auxiliary, which is suffixed onto the main
verb stem. The complex predicate made by this operation inherits the
empathy-locus of the suffixed verb.  The {\em kureru} form of (`give')
can be used as a suffix, to mark {\sc obj} or {\sc obj2} as the
empathy-locus, as can the deictic verb {\em kuru\/} (`come') The use of
the suffix {\em kureta} is shown in (3).

\noindent (3)\\
\footnotesize
\begin{tabular}{llllll}
Hanako wa & Taroo ni & hon o & yonde-kureta. & & \\
& & book & read-gave & & \\
\multicolumn{6}{l}{{\it ``Hanako gave Taroo a favor of reading a book.''}}\\
\multicolumn{6}{l}{EMPATHY=OBJ2=TAROO} \\
\end{tabular}
\normalsize

The suffixation of verbs
such as {\em iku\/} (`go') and the {\em yaru} form of (`give'), mark
{\sc subject} as the empathy-locus, e.g. {\em itta} in (4).

\noindent (4)\\
\footnotesize
\begin{tabular}{llllll}
Hanako wa & Taroo o & tazunete-itta. & & & \\
& & visit-went & & & \\
\multicolumn{6}{l}{{\it ``Hanako went to visit Taroo.''}}\\
\multicolumn{6}{l}{EMPATHY=SUBJ=HANAKO} \\
\end{tabular}
\normalsize

The relevance of speaker's empathy to centering is that a discourse entity
realized as the empathy-locus is more salient, so that the empathy-locus
position is ranked higher on the $\rm{Cf}$. Therefore, we use a ranking
for the $\rm{Cf}$ in
Japanese that incorporates {\sc Empathy} as follows:
\begin{quote}
{\bf Cf Ranking for Japanese} \\
{\sc topic} $>$ {\sc empathy} $>$ {\sc subj} $>$ {\sc obj2} $>$ {\sc obj}
\end{quote}

This ranking is a slight variation of that proposed by
Kameyama\cite{Kameyama86a}.  The centering algorithm works by taking the
arguments of the verb and ordering them according to the Cf ranking
for Japanese given above. In the cases where there are zero pronouns,
there will be multiple possibilities for their interpretation and this
will result in there being a priori several possible Cf
lists\footnote{A discourse entity can simultaneously fulfill multiple
roles. The entity is ranked according to the highest ranked role.}.
These Cf lists are filtered according to the centering rules and
constraints in section
\ref{cent-form}. If there are still multiple possibilities, then the
ordering on transitions applies, and {\sc continuing} interpretations
are preferred.

Many cases of the preference for one interpretation over another
follow directly from the distinction between {\sc continuing} and {\sc
retaining}.

\noindent (5) \\
\footnotesize
\begin{tabular}{llllll}
$\rm{U_{n}}$: & & & & & \\
 Taroo wa & paatii ni & syootai-sareta. & & &\\
 & party  to & invited-was & & &\\
\multicolumn{5}{l}{{\it ``Taroo was invited to the party.''}}&\\
\end{tabular}

\begin{tabular}{|llllll|}
\hline
{\bf Cb:} & TAROO & & & & \\
{\bf Cf:} & [TAROO] & & & & \\ \hline
\end{tabular}

\begin{tabular}{llllll}
$\rm{U_{n+1}}$: & & & & & \\
 0 & Hanako o & totemo &   kiniitta. & & \\
 & &  very-much & was-fond-of & & \\
 \multicolumn{5}{l}{{\it ``He liked Hanako very much.''}} &\\
\end{tabular}

\begin{tabular}{|llllll|}
\hline
{\bf Cb:} & TAROO & & & & \\
{\bf Cf:} & [TAROO, & HANAKO]  & & &  \\
    & subj & obj & & & \\ \hline
\end{tabular}

\begin{tabular}{llllll}
$\rm{U_{n+2}}$: & & & & & \\
Kinoo & 0 & 0 & eiga  ni & sasotta rasii.& \\
yesterday & & & movie to & invite seems & \\
\multicolumn{5}{l}{{\it ``Seemingly he invited her to a movie.''}} &\\
\end{tabular}

\begin{tabular}{|llllll|}
\hline
{\bf Cb:} & TAROO & & & & \\
{\bf Cf1:} & [TAROO, & HANAKO] & CONTINUING & & \\
& subj & obj & & &  \\
{\bf Cf2:} & [HANAKO, & TAROO] & RETAINING & & \\
& subj & obj & & & \\  \hline
\end{tabular}
\normalsize

When the centering algorithm applies in (5) to $\rm{U_{n+2}}$,
constraint 3 says the $\rm{Cb}$ must be the highest ranked element of
Cf($\rm{U_{n+1}}$) realized in $\rm{U_{n+2}}$.  Because there are 2
zeros in $\rm{U_{n+2}}$, TAROO must be realized and therefore must be
the $\rm{Cb}$.  The only {\sc continuing} interpretation available,
{\em Taroo invited Hanako ...}, corresponds to the forward centers
list Cf1.  The fact that the preferred interpretation
is the one in which the {\sc subject} zero pronoun
takes a {\sc subject} antecedent is epiphenomenal.

Example (6) demonstrates the effect of speaker's empathy on
the salience of discourse entities.

\noindent (6) \\
\nopagebreak
\footnotesize
\begin{tabular}{llllll}
$\rm{U_{n}}$: & & & & & \\
Hanako wa & tosyokan de & benkyoositeita. & & &\\
& library  in & studying-was &  &  &\\
\multicolumn{5}{l}{{\it ``Hanako was studying in the library.''}} &\\
\end{tabular}

\begin{tabular}{|llllll|}
\hline
{\bf Cb:} & HANAKO & & & & \\
{\bf Cf:} & [HANAKO] & & & & \\ \hline
\end{tabular}

\footnotesize
\begin{tabular}{llllll}
$\rm{U_{n+1}}$: & & & & & \\
Taroo ga & Hanako o & tetudatte-kureta. & & & \\
& & help-gave  & & & \\
\multicolumn{5}{l}{{\it ``Taroo gave Hanako a favor in helping her.''}} &\\
\end{tabular}

\begin{tabular}{|llllll|}
\hline
{\bf Cb:} & [HANAKO] & & & & \\
{\bf Cf:} & [HANAKO, & TAROO]  & & & \\
    & empathy & subj & & & \\ \hline
\end{tabular}

\begin{tabular}{llllll}
$\rm{U_{n+2}}$: & & & & & \\
Tugi no hi & 0 & 0 & eiga  ni & sasotta. & \\
next of day &SUBJ & OBJ & movie to & invited & \\
\multicolumn{5}{l}{{\it ``Next day she invited him to a movie.''}}& \\
\end{tabular}

\begin{tabular}{|llllll|}
\hline
{\bf Cb:} & HANAKO & & & & \\
{\bf Cf:} & [HANAKO, & TAROO] & CONTINUING & & \\
& subj & obj & & & \\  \hline
\end{tabular}
\normalsize

In (6), HANAKO is the most highly ranked entity from $\rm{U_{n+1}}$
realized in $\rm{U_{n+2}}$ , and therefore must be the $\rm{Cb}$. The
preferred interpretation will therefore be the {\em she invited
him...} one that results from the more highly ranked {\sc continuing}
transition, in which HANAKO is the preferred center ($\rm{Cp}$).

The centering algorithm can also be applied successfully to
intrasentential anaphora, by treating the subordinate clause as though
it were a separate utterance for the purposes of pronoun
interpretation. Consider:

\noindent (7) \\
\footnotesize
\begin{tabular}{llllll}
Taroo wa & Kim ni &[0  0  & bengosuru] &koto o &hanasita. \\
           &  &  & defend &  comp  & told  \\
\multicolumn{6}{l}{{\it``Taroo told Kim that he would defend her''}}\\
\end{tabular}

\begin{tabular}{|llllll|}
\hline
{\bf Cb:} & TAROO & & & & \\
{\bf Cf1:} & [TAROO, & KIM] & CONTINUING & & \\
& subj/top & obj2 & & &  \\
{\bf Cf2:} & [HANAKO, & KIM] & RETAINING & & \\
& subj/top & obj2 & & & \\  \hline
\end{tabular}
\normalsize

The {\sc continuing} interpretation, {\em Taroo told Kim that he
would defend her}, is preferred to the {\sc retaining} interpretation,
{\em Taroo told Kim that she would defend him}.

\section{Topic ambiguity}
The centering process reduces but does not necessarily eliminate
semantic ambiguity in
Japanese discourse. Within a loosely defined context, a native
speaker's intuitions sometimes
still allow for more than one equally preferred interpretation of an utterance.

\subsection{Center Establishment}
In the ``Introduce'' example shown in (8) below, ambiguity
arises from the combined facts that the $\rm{Cb}$ of $\rm{U_{1}}$ is
neutral (undefined),
and there are more entities on the Cf list of $\rm{U_{1}}$ than there are zero
pronouns in $\rm{U_{2}}$.

\noindent (8)\\
\nopagebreak
\footnotesize
\begin{tabular}{llllll}

$\rm{U_{1}}$: & & & & & \\
Lyn-ga & Masayo-ni & Sharon-o & shookaisita & &  \\
SUBJ & OBJ2 & OBJ & introduced & & \\
\multicolumn{6}{l}{\it {``Lyn introduced Sharon to Masayo.''}}\\
\end{tabular}

\begin{tabular}{|llllll|}
\hline
{\bf Cb:} & [?] & & & & \\
{\bf Cf:} & [LYN, & MASAYO, & SHARON] & & \\
   & subj & obj2 & obj & & \\ \hline
\end{tabular}

\begin{tabular}{llllll}

$\rm{U_{2}}$: & & & &  & \\
0 & 0 & kiniitteiru & & & \\
\multicolumn{6}{l}{\it {``Lyn likes Masayo''} (Cf1a)}\\
\multicolumn{6}{l}{\it {``Lyn likes Sharon''} (Cf1b)}\\
\multicolumn{6}{l}{\it {``Masayo likes Sharon''} (Cf2)}\\
\end{tabular}

\begin{tabular}{|llllll|}
\hline
{\bf Cb1:} & LYN & & & & \\
{\bf Cb2:} & MASAYO & & & & \\
{\bf Cf1a:} & [LYN, & MASAYO] & & & \\
      & subj & obj & & & \\
{\bf Cf1b:} & [LYN, & SHARON] & & & \\
      & subj & obj & & &\\
{\bf Cf2:}  & [MASAYO, & SHARON] & & & \\
      & subj & obj & & &\\ \hline
\end{tabular}

\normalsize

All three of these readings of $\rm{U_{2}}$ are equally preferred {\sc
continuations}.  To explain this fact, we posit that the $\rm{Cb}$ of
an initial utterance $\rm{U_{n}}$ may be treated as a variable,
indicated by [?], which can be equated with whatever $\rm{Cb}$ is
assigned to the subsequent utterance $\rm{U_{n+1}}$\footnote{Future
work will discuss center
establishment in more detail, as well as other
interactions, e.g., the effect of {\it wa} marking.}. For example,
because there are 2 zeros in $\rm{U_{2}}$ of (8) and there are 3
entities available to fill these positions, constraint 3 implies that
SHARON (the lowest ranked entity) can never be the $\rm{Cb}$, since it
will never be the most highly ranked element of $\rm{Cf(U_1)}$
realized in $\rm{U_2}$.  Therefore, whenever LYN is realized, the {\sc
continuation} interpretation will place LYN in subject position, thus
explaining the first two readings of $\rm{U_2}$.  The third reading is
available because no $\rm{Cb}$ has yet been established for
$\rm{U_{1}}$, so that a {\sc continuation} does not require the
realization of LYN in $\rm{U_{2}}$. Notice that any reading that
assigns SHARON to the subject position or LYN to a non-subject
position would produce a {\sc retention}.

\subsection{Zero Topics}
Another class of ambiguities can result from the optional assignment of
{\sc topic} to
a zero pronoun.  We propose a topic assignment rule:

\begin{quote}
{\bf Zero Topic Assignment} \\
When no {\sc continuation} transition is available, and a zero pronoun
in $\rm{U_{m}}$ represents an entity that was the $\rm{Cb(U_{m-1})}$
and if no other entity in $\rm{U_{m}}$ is overtly marked as the {\sc
topic}, that zero may be interpreted as the {\sc topic} of
$\rm{U_{m}}$.
\end{quote}

This fact, which has been overlooked in previous treatments of zero
pronouns in Japanese, explains the interesting contrast between the
two discourse segments in examples (9) and (10) below.  Assume in
(9) and (10) that TAROO and HANAKO have already been under
discussion:\footnote{Due to lack of space, we can not discuss the
interaction of center establishment with zero topic assignment here.}

\noindent (9)\\
\footnotesize
\begin{tabular}{llllll}



$\rm{U_{n}}$: & & & & & \\
Taroo wa & kooen o & sanpo-siteita & & & \\
SUBJ & park & walk-around & & & \\
\multicolumn{6}{l}{\it {``Taroo was walking around the park''}}\\
\end{tabular}

\begin{tabular}{|llllll|}
\hline
{\bf Cb:} & TAROO & & & & \\
{\bf Cf:} & [TAROO, & PARK] & & & \\
    & subj & obj & & & \\ \hline
\end{tabular}

\begin{tabular}{llllll}
$\rm{U_{n+1}}$: & & & & & \\
Hanako ga & 0 & yatto & mituketa & & \\
SUBJ &  & finally & found & & \\
\multicolumn{6}{l}{\it {``Hanako finally found (him).''}}\\
\end{tabular}

\begin{tabular}{|llllll|}
\hline
{\bf Cb:} & TAROO & & & & \\
{\bf Cf1:} & [TAROO, & HANAKO] & (C) &  &\\
     & topic/obj & subj & & & \\
{\bf Cf2:} & [HANAKO, & TAROO] & (R) & & \\
     & subj & obj & & & \\ \hline
\end{tabular}

\begin{tabular}{llllll}
$\rm{U_{n+2}}$: & & & & & \\
0 & 0 & yotei o & setumeisita & & \\
SUBJ& OBJ & schedule & explained & & \\
\multicolumn{6}{l}{\it{He explained the schedule to her.}  (Cf1)}\\
\multicolumn{6}{l}{\it{She explained the schedule to him.} (Cf2)}\
\end{tabular}

\begin{tabular}{|llllll|}
\hline
{\bf Cb1:} & TAROO & & & & \\
{\bf Cb2:} & HANAKO & & & & \\
{\bf Cf1:} & [TAROO, & HANAKO] & (C) & & \\
     &  subj & obj & & &\\
{\bf Cf2:} & [HANAKO, & TAROO] & (S-1) & & \\
     &         subj & obj & & &\\ \hline
\end{tabular}

\normalsize

In (9), there are actually two possible Cf lists in $\rm{U_{n+1}}$;
Cf2, which is the only list possible without topic ambiguity,
represents a {\sc retention} (R) rather than a {\sc continuation} (C),
thus triggering zero topic assignment. The utterance $\rm{U_{n+1}}$,
actually has the same meaning for both Cf lists.  The ambiguity in
$\rm{U_{n+2}}$ is caused by the fact that the hearer simultaneously
entertains both of the $\rm{Cf(U_{n+1})}$.  The availability of zero
topic assignment means that TAROO can be the $\rm{Cp}$ even when TAROO
is realized as the topic/object.  The {\sc shift-1} interpretation
results from the algorithm's application to Cf2 of $\rm{U_{n+1}}$.  We
can test to see if topic ambiguity is actually the discourse
phenomenon at work here by contrasting (9) with its minimal pair (10),
in which overt topic marking in $\rm{U_{n+1}}$ rules out topic ambiguity.

\noindent (10)\\
\footnotesize
\begin{tabular}{llllll}
$\rm{U_{n}}$: & & & & & \\
Taroo wa & kooen o & sanpo-siteita & & & \\
SUBJ & park & walk-around & & & \\
\multicolumn{6}{l}{\it {``Taroo was walking around the park''}}\\
\end{tabular}

\begin{tabular}{|llllll|}
\hline
{\bf Cb:} & TAROO & & & & \\
{\bf Cf:} & [TAROO, & PARK] & & & \\
    & subj & obj & & & \\ \hline
\end{tabular}

\begin{tabular}{llllll}
$\rm{U_{n+1}}$: & & & & & \\
Hanako-wa & 0 & yatto & mituketa & & \\
TOP/SUBJ &  & finally & found & & \\
 \multicolumn{6}{l}{\it {``Hanako finally found (him).''}}\\
\end{tabular}

\begin{tabular}{|llllll|}
\hline
{\bf Cb:} & TAROO & & & & \\
{\bf Cf:} & [HANAKO, & TAROO] & (R) & & \\
    & top/subj & obj & & & \\ \hline
\end{tabular}

\begin{tabular}{llllll}
$\rm{U_{n+2}}$: & & & & & \\
0 & 0 & yotei-o & setumeisita & & \\
&   & schedule & explained & & \\
\multicolumn{6}{l}{\it {``She explained the schedule to him''}}\\
\end{tabular}

\begin{tabular}{|llllll|}
\hline
{\bf Cb:} & HANAKO & & & & \\
{\bf Cf:} & [HANAKO, &  TAROO] &(SHIFT1) & & \\
    & subj & obj & & & \\ \hline
\end{tabular}

\normalsize

In (10) the only Cf possible for $\rm{U_{n+1}}$ is the {\sc retention}
in the parallel utterance in (9). Given that there are 2 zero
pronouns in $\rm{U_{n+2}}$, constraint 3 forces a shift.  The {\em
Hanako explained ...} interpretation is preferred because it is the
more highly ranked {\sc shift-1} transition.  If {\sc hanako} could
represent a {\sc topic-obj} there would be another equally ranked {\sc
shift-1} interpretation.  However, HANAKO can not be a zero topic
because it was not the $\rm{Cb}$ of the previous utterance.

\section{Discussion}

We have demonstrated a computational treatment of the resolution of
zero pronouns in Japanese. Kameyama proposed an analysis of Japanese
zero pronouns that used centering, but did not distinguish between
{\sc continuing} and {\sc retaining}, and thus required an extra
mechanism, i.e. property-sharing\cite{Kameyama85}.  Our examples (5), (6)
and (7) show that property-sharing is an unnecessary stipulation.  In
addition, there are a number of cases in which property-sharing just
doesn't work. Our ``introduce'' example (8) illustrates that it is not
essential for a zero pronoun to share a grammatical function property
with its antecedent. In fact property-sharing would falsely predict that the
{\em Masayo likes Sharon} interpretation of (8) $\rm{U_2}$ is not
possible, as well as falsely predicting the ungrammaticality of
examples like (11) below.

\noindent(11)\\
\footnotesize
\begin{tabular}{llllll}
$\rm{U_{n}}$: & & & & & \\
 & Hanako wa &  repooto o & kaita. & & \\
&  & report & wrote & & \\
& \multicolumn{5}{l}{\it {``Hanako wrote a report''}}\\
& & & & & \\
$\rm{U_{n+1}}$: & & & & & \\
& $0_{i}$ & Taroo ni & aini-itta. & & \\
& & & to see-went & & \\
& \multicolumn{5}{l}{\it {``She went to see Taroo''}}\\
& \multicolumn{5}{l}{$0_{i}$ = Hanako [SUB EMPATHY]}\\
& & & & & \\
$\rm{U_{n+2}}$: & & & & & \\
& Taroo wa & $0_{i}$ & kibisiku hihansita. & & \\
& & & severely criticized & & \\
& \multicolumn{5}{l}{\it {``Taroo severely criticized her.''}}\\
& \multicolumn{5}{l}{$0_{i}$ = Hanako [nonSUB nonEMPATHY]}\\
\end{tabular}
\normalsize

Property-sharing requires that in $\rm{U_{n+2}}$, $i \neq $ HANAKO,
since the zero  carries the properties ({\sc subj},
{\sc empathy}) in $\rm{U_{n+1}}$, but has
the properties ({\sc nonsubj},{\sc nonempathy}) in
$\rm{U_{n+2}}$\footnote{Kameyama called the Empathy property IDENT.}.
But in fact $\rm{U_{n+2}}$ is perfectly acceptable under the intended
reading of {\em Taroo severely criticized Hanako}.  Nothing special
needs to be said about these to get the correct interpretation using
the centering algorithm.

We have also proposed a notion of topic ambiguity, which arises from
the fact that the grammatical function of unexpressed zero arguments
is indeterminate. The application of zero topic assignment also
depends on the centering theory distinction between {\sc continuing}
and {\sc retaining}.  In addition, the centering construct of
backward-looking center, $\rm{Cb}$, gives us a computational way of
determining when a zero pronoun may be assigned {\sc Topic}.  Topic
ambiguity has been ignored in previous analyses, but it explains the
availability of interpretations that previous accounts would predict
as ungrammatical.

This analysis has implications for the design of language-independent
discourse processing modules. We claim that the syntactic factors that
affect the ranking of the items on the forward center list, $\rm{Cf}$,
will vary from language to language. The ordering for Japanese
incorporates {\sc topic} and {\sc empathy} into the $\rm{Cf}$ ranking,
which is a single parameter of the centering algorithm.  In every
other respect, the rules and constraints of the centering framework
that the centering algorithm implements remain invariant.

\section{Acknowledgements}

The authors would like to thank Aravind Joshi, Carl Pollard, and Ellen
Prince for their comments and support. This paper has also benefited
from suggestions by Megan Moser, Peter Sells, Enric Vallduv\'{\i},
Bonnie Webber and Steve Whittaker.  This research was partially funded
by ARO grants DAAG29-84-K-0061 and DAAL03-89-C0031PRI, DARPA grant
N00014-85-K0018, and NSF grant MCS-82-19196 at the University of
Pennsylvania, and by Hewlett-Packard Laboratories.



\begin{thebibliography}{GJW86}

\bibitem[BF83]{BF83}
Roger Brown and Deborah Fish.
\newblock The psychological causality implicit in language.
\newblock {\em Cognition}, 14:237--273, 1983.

\bibitem[BFP87]{BFP87}
Susan~E. Brennan, Marilyn~Walker Friedman, and Carl~J. Pollard.
\newblock A centering approach to pronouns.
\newblock In {\em Proc. 25th Annual Meeting of the ACL, Stanford}, pages
  155--162, 1987.

\bibitem[Bre89]{Brennan89}
Susan~E. Brennan.
\newblock Centering attention in discourse. Unpublished manuscript.
\newblock 1989.

\bibitem[GJW83]{GJW83}
Barbara~J. Grosz, Aravind~K. Joshi, and Scott Weinstein.
\newblock Providing a unified account of definite noun phrases in discourse.
\newblock In {\em Proc. 21st Annual Meeting of the ACL, Association of
  Computational Linguistics}, pages 44--50, 1983.

\bibitem[GJW86]{GJW86}
Barbara~J. Grosz, Aravind~K. Joshi, and Scott Weinstein.
\newblock Towards a computational theory of discourse interpretation.
\newblock Unpublished Manuscript, 1986.

\bibitem[Gri75]{Grice75}
H.P. Grice.
\newblock Logic and conversation.
\newblock In P.~Cole and J.~Morgan, editors, {\em Syntax and Semantics III -
  Speech Acts}, pages 41--58. Academic Press, New York, 1975.

\bibitem[Gro77]{Grosz77}
Barbara~J. Grosz.
\newblock The representation and use of focus in dialogue understanding.
\newblock Technical Report 151, SRI International, 333 Ravenswood Ave, Menlo
  Park, Ca. 94025, 1977.

\bibitem[GS85]{GS85}
Barbara~J. Grosz and Candace~L. Sidner.
\newblock The structure of discourse structure.
\newblock Technical Report CSLI-85-39, Center for the Study of Language and
  Information, Stanford, CA, 1985.

\bibitem[HD88]{Hudson88}
Susan~B. Hudson-D'Zmura.
\newblock {\em The Structure of Discourse and Anaphor Resolution: The discourse
  Center and the Roles of Nouns and Pronouns}.
\newblock PhD thesis, University of Rochester, 1988.

\bibitem[JW81]{JW81}
Aravind~K. Joshi and Scott Weinstein.
\newblock Control of inference: Role of some aspects of discourse structure -
  centering.
\newblock In {\em Proc. International Joint Conference on Artificial
  Intelligence}, pages 385--387, 1981.

\bibitem[Kam85]{Kameyama85}
Megumi Kameyama.
\newblock {\em Zero anaphora: the case of Japanese}.
\newblock PhD thesis, Stanford University, 1985.

\bibitem[Kam88]{Kameyama86a}
Megumi Kameyama.
\newblock Japanese zero pronominal binding, where syntax and discourse meet.
\newblock In William Poser, editor, {\em Papers from the Second International
  Workshop on Japanese Syntax}, pages 47--74. Stanford: CSLI, 1988.
\newblock also available as University of Pennsylvania Tech Report
  MS-CIS-86-60.

\bibitem[KK77]{Kuno-Kab77}
Susumu Kuno and Etsuko Kaburaki.
\newblock Empathy and syntax.
\newblock {\em Linguistic Inquiry}, 8:627--672, 1977.

\bibitem[Kun73]{Kuno73}
Susumo Kuno.
\newblock {\em The Structure of the Japanese Language}.
\newblock The MIT Press, Cambridge, Massachusetts, 1973.

\bibitem[Pri81]{Prince81}
Ellen~F. Prince.
\newblock Toward a taxonomy of given-new information.
\newblock In {\em Radical Pragmatics}, pages 223--255. Academic Press, 1981.

\bibitem[Pri85]{Prince85}
Ellen~F. Prince.
\newblock Fancy syntax and shared knowledge.
\newblock {\em Journal of Pragmatics}, pages 65--81, 1985.

\bibitem[Sid79]{Sidner79}
Candace~L. Sidner.
\newblock Toward a computational theory of definite anaphora comprehension in
  {English}.
\newblock Technical Report AI-TR-537, MIT, 1979.

\bibitem[SSJ74]{SSJ74}
Harvey Sacks, Emmanuel Schegloff, and Gail Jefferson.
\newblock A simplest systematics for the organization of turn-taking in
  conversation.
\newblock {\em Language}, 50:325--345, 1974.

\bibitem[Wal89]{Walker89b}
Marilyn~A. Walker.
\newblock Evaluating discourse processing algorithms.
\newblock In {\em Proc. 27th Annual Meeting of the Association of Computational
  Linguistics}, pages 251--261, 1989.

\end{thebibliography}
\end{document}